\documentclass[10pt,twocolumn,letterpaper]{article}

\usepackage{cvpr}
\usepackage{times}
\usepackage{epsfig}
\usepackage{graphicx}
\usepackage{amsmath}
\usepackage{amssymb}

\usepackage[pagebackref=true,breaklinks=true,letterpaper=true,colorlinks,bookmarks=false]{hyperref}

\cvprfinalcopy % *** Uncomment this line for the final submission

\def\cvprPaperID{728} % *** Enter the CVPR Paper ID here

\ifcvprfinal\fi
\begin{document}

%%%%%%%%% TITLE
\title{Deep Matching Prior Network: Toward Tighter Multi-oriented Text Detection}

\author{Yuliang Liu, Lianwen Jin+\\
College of Electronic Information Engineering\\
South China University of Technology \\
% Guangzhou China\\
{\tt\small +lianwen.jin@gmail.com}
% \author{Yuliang Liu\\
% \and
}

\maketitle
%\thispagestyle{empty}

%%%%%%%%% ABSTRACT
\begin{abstract}
   % Detecting incidental scene text is challenging because of multi-orientation, perspective distortion, and variation of text size, color and scale. Previous researches only focus on using rectangular bounding box or horizontal sliding window to localize or recall text, which may result in redundant background noise, unnecessary overlap and even information loss.
   % From these issues, in this paper, we propose a CNN based system, named deep matching prior network (DMPNet), to detect text with tighter quadrangle. First we use quadrilateral sliding windows in several intermediate convolutional layers, to roughly recall text. After that, we use a proposed strategy to finely predict text with tight quadrangle based on our sequential protocol.
   % In addition, due to the previous protocol of computing overlap can not compute polygonal overlapping area and can not provide satisfactory computational accuracy, we proposed a shared Monte-Carlo method that has both high speed and accuracy properties in computing the polygonal area.
   % Moreover, we also proposed a smooth ln loss for further finely regressing the position of text, which has better overall performance than L2 loss and smooth L1 loss in terms of robustness and stability.
   % The effectiveness of our system is evaluated on public word-level, multi-oriented and most challenging scene text database, ICDAR 2015 Robust Reading Competition Challenge 4 "Incidental scene text". The f measure of our system is 70.64 percent much better than the state-of-the-art performance.
   Detecting incidental scene text is a challenging task because of multi-orientation, perspective distortion, and variation of text size, color and scale. Retrospective research has only focused on using rectangular bounding box or horizontal sliding window to localize text, which may result in redundant background noise, unnecessary overlap or even information loss. To address these issues, we propose a new Convolutional Neural Networks (CNNs) based method, named Deep Matching Prior Network (DMPNet), to detect text with tighter quadrangle. First, we use quadrilateral sliding windows in several specific intermediate convolutional layers to roughly recall the text with higher overlapping area and then a shared Monte-Carlo method is proposed for fast and accurate computing of the polygonal areas. After that, we designed a sequential protocol for relative regression which can exactly predict text with compact quadrangle. Moreover, a auxiliary smooth Ln loss is also proposed for further regressing the position of text, which has better overall performance than L2 loss and smooth L1 loss in terms of robustness and stability. The effectiveness of our approach is evaluated on a public word-level, multi-oriented scene text database, ICDAR 2015 Robust Reading Competition Challenge 4 ``Incidental scene text localization''. The performance of our method is evaluated by using F-measure and found to be 70.64\%, outperforming the existing state-of-the-art method with F-measure 63.76\%.

   % Online demo is in ... .
   % (if needed) We also release second version of SCUT-FORLU multi-orientation scene text data set, which is available at http://
   % This method detect word

\end{abstract}

%%%%%%%%% BODY TEXT
\section{Introduction}

Scene text detection is an important prerequisite~\cite{Weinman2009Scene,Weinman2014Toward,Yin2014Robust,Bissacco2013PhotoOCR,Ye2015Text} for many content-based  applications, e.g., multilingual translation, blind navigation and automotive assistance. Especially, the recognition stage always stands in need of localizing scene text in advance, thus it is a significant requirement for detecting methods that can tightly and robustly localize scene text.

\begin{figure}[htb]
% \begin{minipage}[c]{1\linewidth}
%   \centering
%   \centerline{\includegraphics[width = 8.0cm]{74.eps}}
% %  \vspace{1.5cm}
%   \centerline{(a) forward loss }\medskip
% \end{minipage}
% \vfill  % hfill horizontal  vfill vertical
\begin{minipage}[c]{01\linewidth}
  \centering
  \centerline{\includegraphics[width = 8.0cm, height = 2.0cm]{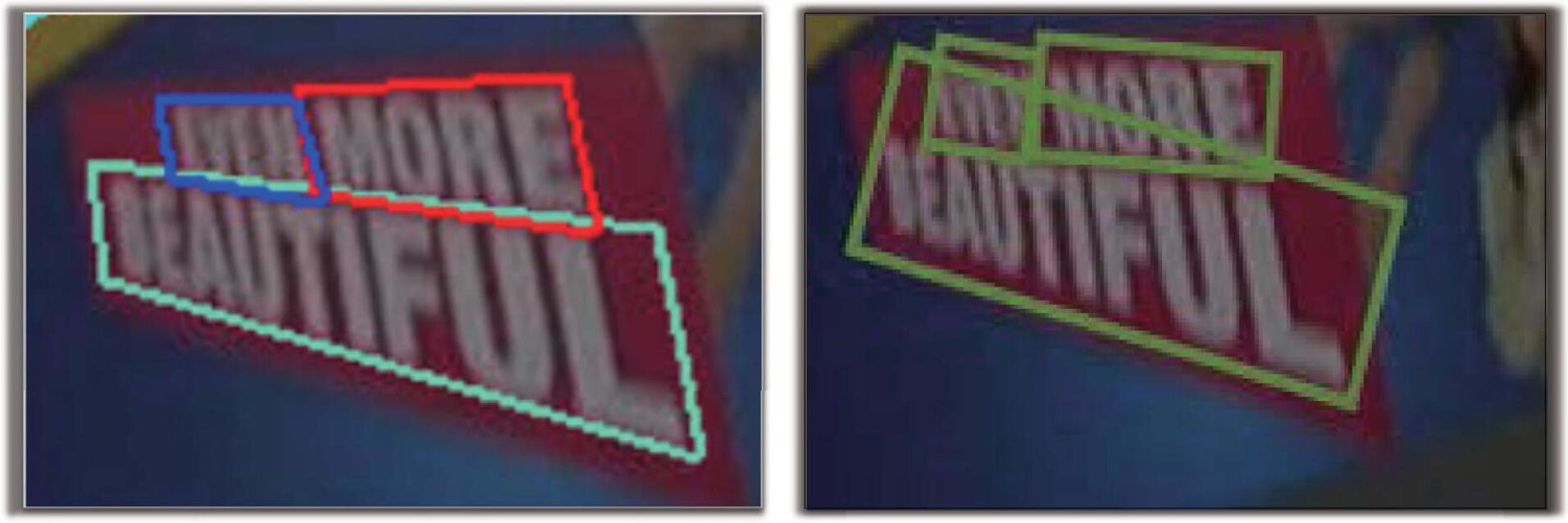}}
%  \vspace{1.5cm}
  \centerline{\small{(a) Rectangular bounding box cause unnecessary overlap. }}\medskip
\end{minipage}
% \vfill  % hfill horizontal  vfill vertical
% \begin{minipage}[c]{01\linewidth}
%   \centering
%   \centerline{\includegraphics[width = 8.0cm, height = 2.6cm]{390.eps}}
% %  \vspace{1.5cm}
%   \centerline{\small{(b) Redundant background noise.}}\medskip
% \end{minipage}
\vfill  % hfill horizontal  vfill vertical
\begin{minipage}[c]{01\linewidth}
  \centering
  \centerline{\includegraphics[width = 8.0cm, height = 3.5cm]{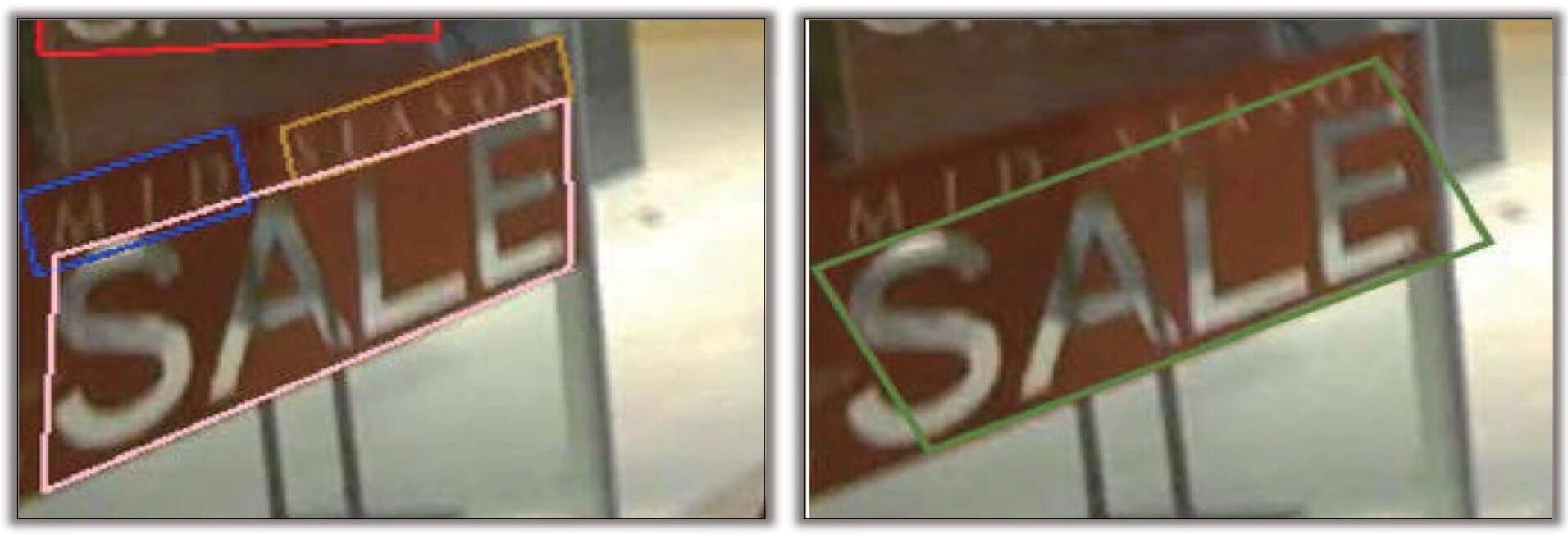}}
%  \vspace{1.5cm}
  \centerline{\small{(c) Marginal text can not be exactly localized with rectangle. }}\medskip
\end{minipage}
\vfill  % hfill horizontal  vfill vertical
\begin{minipage}[c]{01\linewidth}
  \centering
  \centerline{\includegraphics[width = 8.0cm, height = 1.7cm]{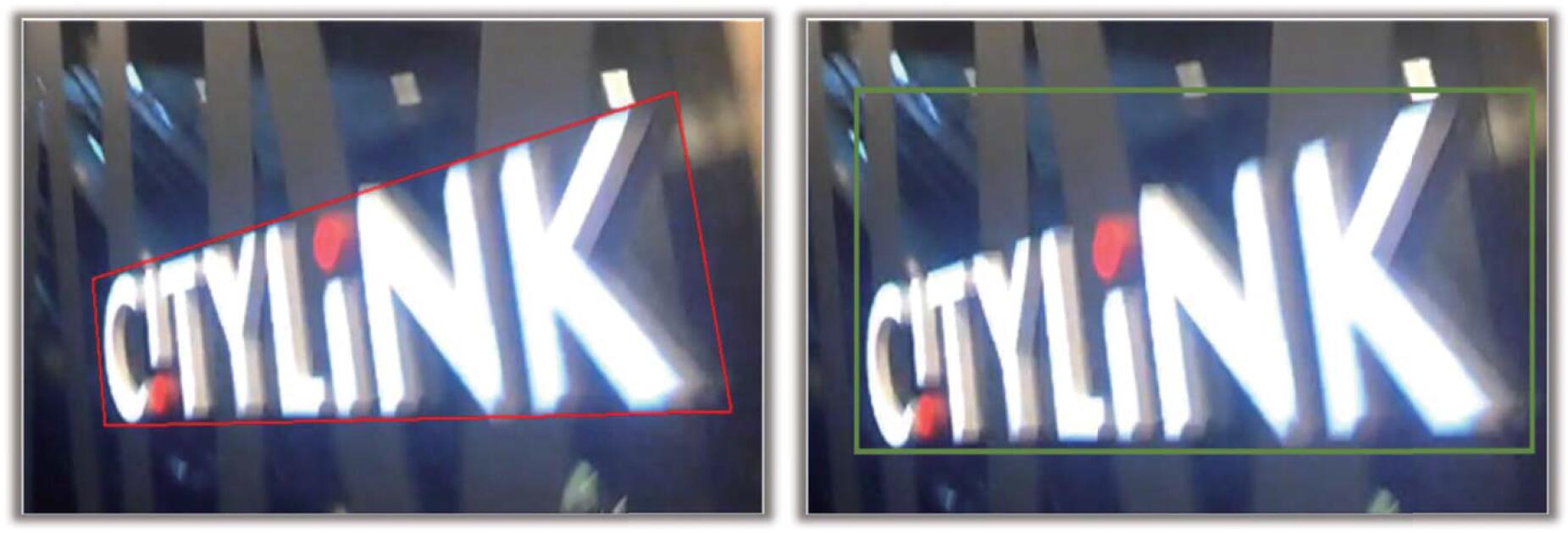}}
%  \vspace{1.5cm}
  \centerline{\small{(d) Rectangular bounding box brings redundant noise. }}\medskip
\end{minipage}
% \vfill  % hfill horizontal  vfill vertical
% \begin{minipage}[c]{01\linewidth}
%   \centering
%   \centerline{\includegraphics[width = 8.0cm, height = 1.8cm]{843.eps}}
% %  \vspace{1.5cm}
%   \centerline{\small{(e) Redundant background noise. }}\medskip
% \end{minipage}
%
\caption{Comparison of quadrilateral bounding box and rectangular bounding box for localizing texts.}\label{fig:background}
\end{figure}

Camera captured scene text are often found with low-quality; these texts may have multiple orientations, perspective distortions, and variation of text size, color or scale~\cite{Zhu2016Scene}, which makes it a very challenging task~\cite{zhang2016multi}.
 In the past few years, various existing methods have successfully been used for detecting horizontal or near-horizontal texts~\cite{Chen2004Detecting,Epshtein2010Detecting,Neumann2013Scene,jaderberg2014deep,Huang2014Robust}. However, due to the horizontal rectangular constraints, multi-oriented text are restrictive to be recalled in practice,~\eg low accuracies reported in ICDAR 2015 Competition Challenge 4 ``Incidental scene text localization''~\cite{Karatzas2015ICDAR}.

Recently, numerous techniques~\cite{Yi2011Text,Yin2015Multi,Kang2014Orientation,zhang2016multi} have been devised for multi-oriented text detection; these methods used rotated rectangle to localize oriented text. However, Ye and Doermann~\cite{Ye2015Text} indicated that because of characters distortion, the boundary of text may lose rectangular shape, and the rectangular constraints may result in redundant background noise, unnecessary overlap or even information loss when detecting distorted incidental scene text as shown in Figure~\ref{fig:background}.
It can be visualized from the Figure that the rectangle based methods must face three kinds of circumstances: i) redundant information may reduce the reliability of detected confidence~\cite{Li2006Confidence} and make subsequent recognition harder~\cite{Zhu2016Scene}; ii) marginal text may not be localized completely; iii) when using non-maximum suppression~\cite{Neubeck2006Efficient}, unnecessary overlap may eliminate true prediction.

To address these issues, in this paper, we proposed a new Convolutional Neural Networks (CNNs) based method, named Deep Matching Prior Network (DMPNet), toward tighter text detection. To the best of our knowledge, this is the first attempt to detect text with quadrangle. Basically, our method consists of two steps: roughly recall text and finely adjust the predicted bounding box. First, based on the priori knowledge of textual intrinsic shape, we design different kinds of quadrilateral sliding windows in specific intermediate convolutional layers to roughly recall text by comparing the overlapping area with a predefined threshold. During this rough procedure, because numerous polygonal overlapping areas between the sliding window (SW) and ground truth (GT) need to be computed, we design a shared Monte-Carlo method to solve this issue, which is qualitatively proved more accurate than the previous computational method~\cite{Tu2012Detecting}.
% It is observed that the previous methods~\cite{Tu2012Detecting} can only compute rectangular area with unsatisfactory computational accuracy, we proposed a shared Monte-Carlo method that has both high speed and accuracy properties for computing the polygonal area.
% Kang2014Orientation Authors14
After roughly recalling text, those SWs with higher overlapping area would be finely adjusted for better localizing; different from existing methods~\cite{Chen2004Detecting,Epshtein2010Detecting,Neumann2013Scene,jaderberg2014deep,Huang2014Robust,Yi2011Text,Yin2015Multi,zhang2016multi} that predict text with rectangle, our method can use quadrangle for tighter localizing scene text, which owe to the sequential protocol we purposed and the relative regression we used.
 % we use a new strategy to finely predict text with tight quadrangle based on a sequential protocol, we proposed.
Moreover, a new smooth $Ln$ loss is also proposed for further regressing the position of text, which has better overall performance than $L2$ loss and smooth $L1$ loss in terms of robustness and stability. Experiments on the public word-level and multi-oriented dataset, ICDAR 2015 Robust Reading Competition Challenge 4 ``Incidental scene text localization'', demonstrate that our method outperforms previous state-of-the-art methods~\cite{Yao2015Incidental} in terms of F-measure.

We summarize our contributions as follow:
\begin{itemize}
  \item  We are the first to put forward prior quadrilateral sliding window, which significantly improve the recall rate.
  \item  We proposed sequential protocol for uniquely determining the order of 4 points in arbitrary plane convex quadrangle, which enable our method for using relative regression to predict quadrilateral bounding box.
  \item  The proposed shared Monte-Carlo computational method can fast and accurately compute the polygonal overlapping area.
  \item  The proposed smooth $Ln$ loss has better overall performance than $L2$ loss and smooth $L1$ loss in terms of robustness and stability.
  % \item  DMPNet is a one stage single CNN based method, doesn't need to learn the region proposal proposed in [faster r-cnn]. Thus, our method is fast, with what fps to test one image in an image.
  \item  Our approach shows state-of-the-art performance in detecting incidental scene text.
\end{itemize}

% The prefer situation is that the bounding box can follow the text direction.
% ~\cite{Ye2015Text} pointed out Domain knowledge of text would be future work.
% Some drawback of existing deep learning methods are that they ignores the text feature, simply use successful deep learning models~\cite{Zhu2016Scene} that perform well in pascal VOC [refer] and so on.
% SWT and MSER are two main feature of text, we find the text direction is also a important feature of text, and tightly localizing the ground truth can easily use this feature because the boundaries of the quadrangle represents the text directions as shown in the figure~\ref{fig:background}. This prior domain knowledge is good for our deep learning model. (Explained why our model called deep matching prior network.)

% refer~\cite{Liu2015SSD} detection all based on Faster R-CNN~\cite{Ren2016Faster} albeit with deeper features such as [3]. Although accurate, these approaches have been too computationally intensive for embedded systems and, even with high-end hardware, too slow for real-time or near real-time applications. Often detection speed for these approaches is measured in seconds per frame, and even the fastest high-accuracy detector, the basic Faster R-CNN, operates at only 7 frames per second (FPS). There have been a wide range of attempts to build faster detectors by attacking each stage of the detection pipeline (see related work in Sec. 4), but so far, significantly increased speed comes only at the cost of significantly decreased detection accuracy
% Chen2011Robust
\section{Related work}
Reading text in the wild have been extensively studied in recent years because scene text conveys numerous valuable information that can be used on many intelligent applications,~\eg autonomous vehicles and blind navigation. Unlike generic objects, scene text has unconstrained lengths, shape and especially perspective distortions, which make text detection hard to simply adopt techniques from other domains. Therefore, the mainstream of text detection methods always focused on the structure of individual characters and the relationships between characters~\cite{Zhu2016Scene},~\eg connected component based methods~\cite{zamberletti2014text,Shi2013Scene,Neumann2012Real}. These methods often use stroke width transform (SWT)~\cite{Huang2013Text} or maximally stable extremal region (MSER)~\cite{Matas2004Robust,Nist2008Linear} to first extract character candidates, and using a series of subsequence steps to eliminate non-text noise for exactly connecting the candidates. Although accurate, such methods are somewhat limited to preserve various true characters in practice~\cite{cho2016canny}.
% In this section, we only focus on the most relevant works that are presented
% for multi-oriented text detection. There are a variety of text localization
% techniques in the literature. The most common approach involves three key
% components [36]: character candidate extraction, character classification,
% and text grouping. Grouping text as a set of words or sentences depends on
% the objective of the algorithm and may involve text line estimation and
% validation. Connected component based methods focused on the that utilizes
% character candidates extracted with particular constraints, e.g., consistent
% stroke width or extremal region. (2) sliding window based methods that
% exhaustively scan windows at all possible locations and scales, and

Another mainstream method is based on sliding window~\cite{Chen2004Detecting,Kim2003Texture,hanif2009Text,lee2011adaboost}, which shifts a window in each position with multiple scales from an image to detect text. Although this method can effectively recall text, the classification of the locations can be sensitive to false positives because the sliding windows often carry various background noise.
% The essential properties of scene text lie in the structure of individual characters and the relationships between characters.

Recently, Convolutional Neural Networks~\cite{simonyan2014very,girshick2015fast,Ren2016Faster,liu2015ssd,redmon2015you} have been proved powerful enough to suppress false positives, which enlightened researchers in the area of scene text detection; in~\cite{Huang2014Robust}, Huang~\etal integrated MSER and CNN to significantly enhance performance over conventional methods; Zhang~\etal utilized Fully Convolutional Network~\cite{zhang2016multi} to efficiently generate a pixel-wise text/non-text salient map, which achieve state-of-the-art performance on public datasets.
It is worth mentioning that the common ground of these successful methods is to utilized textual intrinsic information for training the CNN. Inspired by this promising idea, instead of using constrained rectangle, we designed numerous quadrilateral sliding windows based on the textual intrinsic shape, which significantly improves recall rate in practice.

\begin{figure*}[ht]
\begin{minipage}[b]{0.4\linewidth}
  % \centering
  \centerline{\includegraphics[width = 5.0cm]{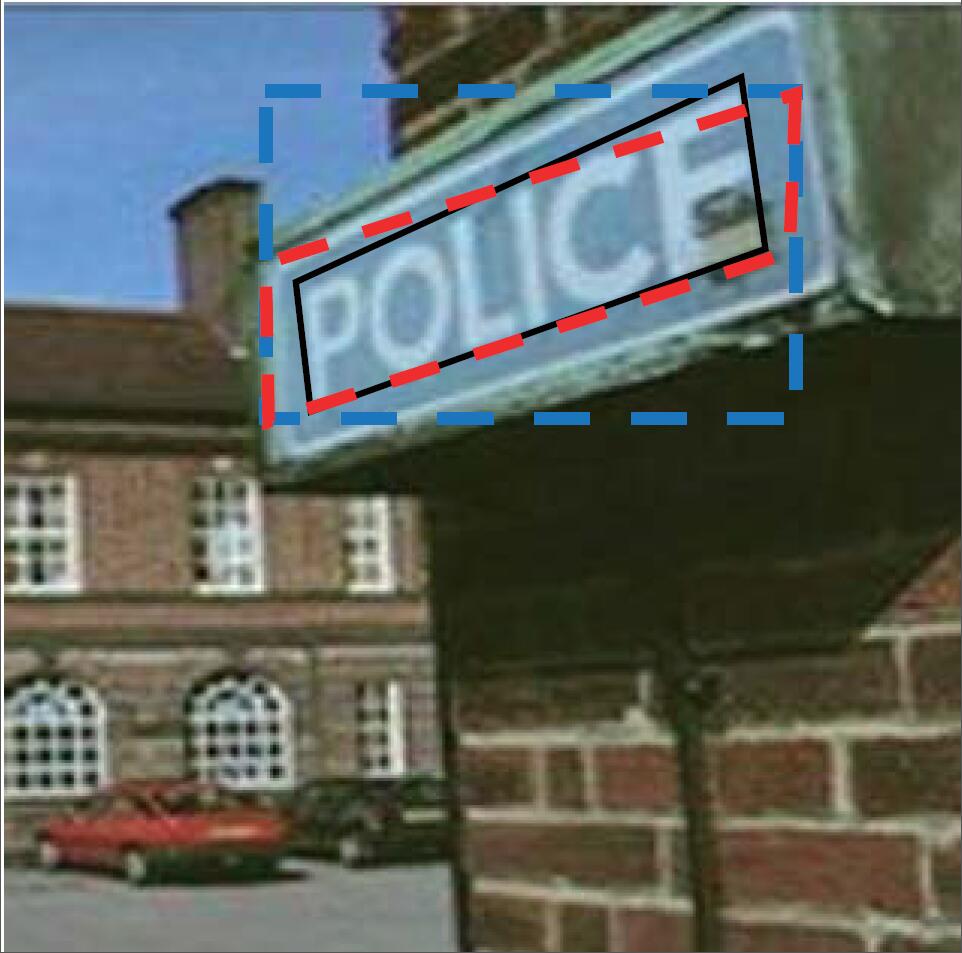}}
%  \vspace{1.5cm}
  \centerline{(a) Comparison of recalling scene text. }
\end{minipage}
\hfill
\begin{minipage}[b]{-0.35\linewidth}
  % \centering
  \centerline{\includegraphics[width = 5.0cm]{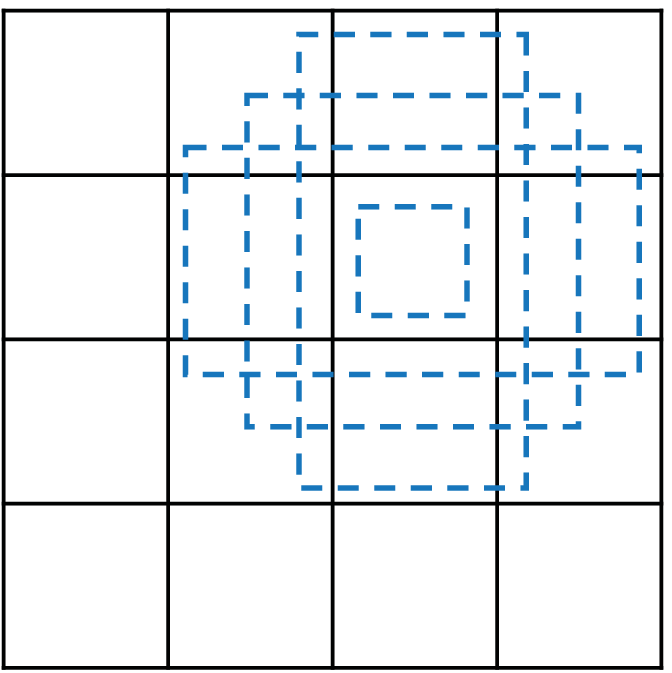}}
%  \vspace{1.5cm}
  \centerline{(b) Horizontal sliding windows. }
\end{minipage}
\hfill
\begin{minipage}[b]{-0.3\linewidth}
  % \centering
  \centerline{\includegraphics[width = 5.0cm]{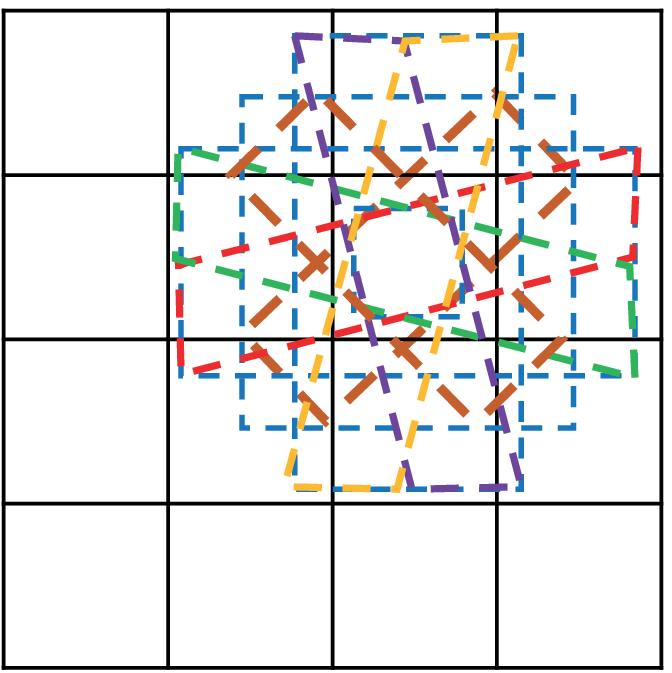}}
%  \vspace{1.5cm}
  \centerline{(c) Proposed quadrilateral sliding windows. }
\end{minipage}
\caption{Comparison between horizontal sliding window and quadrilateral sliding window. (a): Black bounding box represents ground truth; red represents our method. Blue represents horizontal sliding window. It can be visualized that quadrilateral window can easier recall text than rectangular window with higher overlapping area. (b): Horizontal sliding windows used in~\cite{liu2015ssd}. (c): Proposed quadrilateral sliding windows. Different quadrilateral sliding window can be distinguished with different color.}\label{fig:slide}
\end{figure*}

\section{Proposed methodology}
This section presents details of the Deep Matching Prior Network (DMPNet). It includes the key contributions that make our method reliable and accurate for text localization: firstly, roughly recalling text with quadrilateral sliding window; then, using a shared Monte-Carlo method for fast and accurate computing of polygonal areas; finely localizing text with quadrangle and design a Smooth $Ln$ loss for moderately adjusting the predicted bounding box.

\subsection{Roughly recall text with quadrilateral sliding window}
% [cvpr 201604]
% In the past few years, early researchers [] manually designed features to identify characters with strong classifier. Recently, some work [1,2] have achieved great performance, adopting CNN as a character detector. However, even the state-of-the-art character detector [] still performs poorly at complicated background
Previous approaches~\cite{liu2015ssd,Ren2016Faster} have successfully adopted sliding windows in the intermediate convolutional layers to roughly recall text. Although the methods~\cite{Ren2016Faster} can accurately learn region proposal based on the sliding windows, these approaches have been too slow for real-time or near real-time applications. To raise the speed, Liu~\cite{liu2015ssd} simply evaluate a small set of prior windows of different aspect ratios at each location in several feature maps with different scales, which can successfully detect both small and big objects. However, the horizontal sliding windows are often hard to recall multi-oriented scene text in our practice. Inspired by the recent successful methods~\cite{Huang2014Robust,zhang2016multi} that integrated the textual feature and CNN, we put forward numerous quadrilateral sliding windows based on the textual intrinsic shape to roughly recall text.

During this rough procedure, an overlapping threshold was used to judge whether the sliding window is positive or negative. If a sliding window is positive, it would be used to finely localize the text. Basically, a small threshold may bring a lot of background noise, reducing the precision, while a large threshold may make text harder to be recalled. But if we use quadrilateral sliding window, the overlapping area between sliding window and ground truth can be larger enough to reach a higher threshold, which are beneficial to improve both the recall rate and the precision as shown in Figure~\ref{fig:slide}.
% It is observed that various of scene texts are inclined and such inclination would be magnified in the long text as shown in the figure~\ref{fig:background} (a).
As the figure presents, we reserve the horizontal sliding windows, simultaneously designing several quadrangles inside them based on the prior knowledge of textual intrinsic shape: a) two rectangles with 45 degrees are added inside the square; b) two long parallelograms are added inside the long rectangle. c) two tall parallelograms are added inside the tall rectangle.

With these flexible sliding windows, the rough bounding boxes become more accurate and thus the subsequence finely procedure can be easier to localize text tightly.
% the overlapping area between sliding window and ground truth can be larger enough to reach a higher threshold, which are beneficial to improve both the recall rate and the precision.
In addition, because of less background noise, the confidence of these quadrilateral sliding windows can be more reliable in practice, which can be used to eliminate false positives.

% This idea is designed based on the intrinsic feature
% In the past few years, early
% Inspired by [ssd], during the training procedure, we also evaluate a small set of prior boxes of different aspect ratios at each location in several feature maps with different scales (e.g. 25x25 and 12x12 ). Separately, we use. As showed in figure~\ref{fig:slide}

% Because we design these windows for adapting text features, we named it prior window [intro need explain.]
\begin{figure*}[htb]
  \centering
  \centerline{\includegraphics[width=17cm, height = 7.51cm]{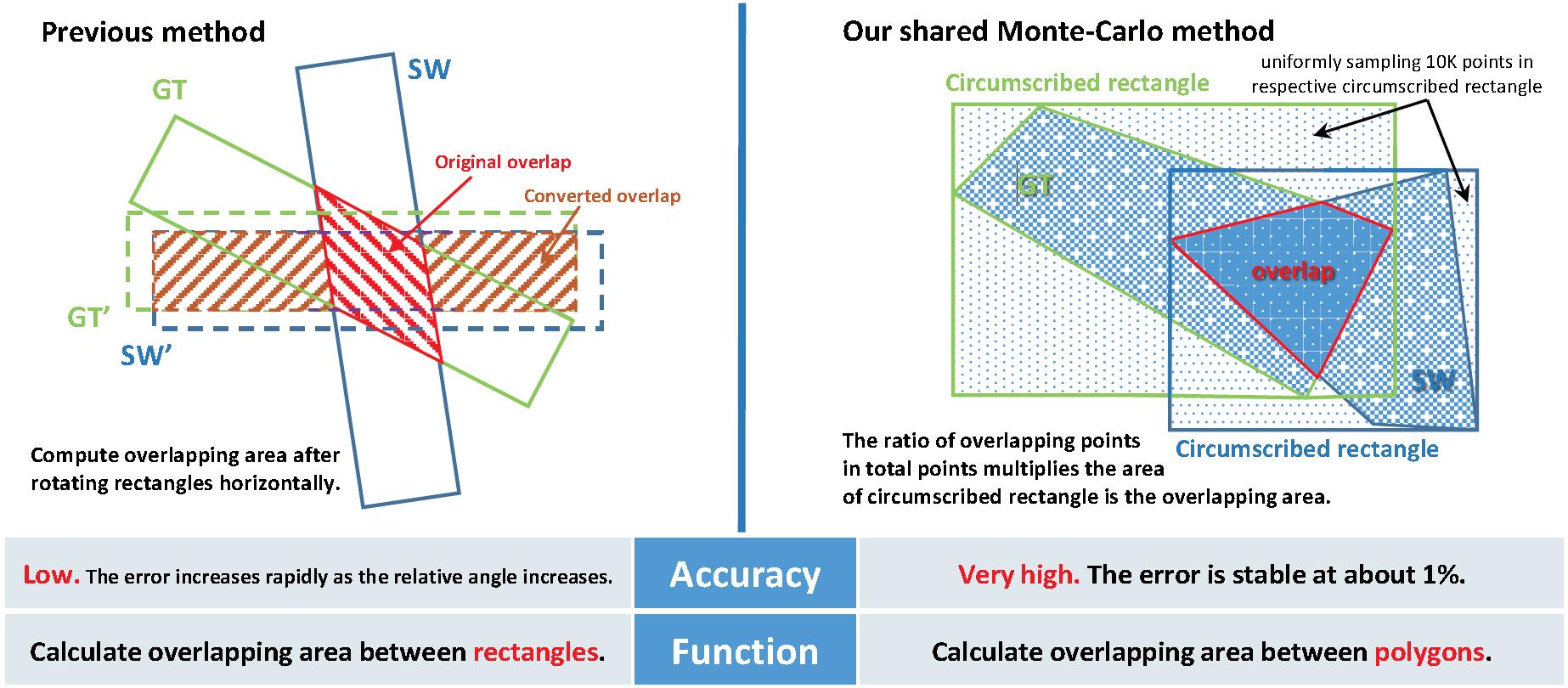}}
%  \vspace{2.0cm}
  \caption{Comparison between previous method and our method in computing overlapping area.}\label{fig:monte}
\end{figure*}

\subsubsection{Shared Monte-Carlo method}
As mentioned earlier, for each ground truth, we need to compute its overlapping area with every quadrilateral sliding window. However, the previous method~\cite{Tu2012Detecting} can only compute rectangular area with unsatisfactory computational accuracy, thus we proposed a shared Monte-Carlo method that has both high speed and accuracy properties when computing the polygonal area. Our method consists of two steps.

a) First, we uniformly sample 10,000 points in circumscribed rectangle of the ground truth. The area of ground truth ($S_{GT}$) can be computed by calculating the ratio of overlapping points in total points multiplied by the area of circumscribed rectangle. In this step, all points inside the ground truth would be reserved for sharing computation.

b) Second, if the circumscribed rectangle of each sliding window and the circumscribed rectangle of each ground truth do not have a intersection, the overlapping area is considered zero and we do not need to further compute.
If the overlapping area is not zero, we use the same sampling strategy to compute the area of sliding window ($S_{SW}$) and then calculating how many the reserved points from the first step inside the sliding window. The ratio of inside points multiplies the area of the circumscribed rectangle is the overlapping area.
Specially, this step is suitable for using GPU parallelization, because we can use each thread to be responsible for calculating each sliding window with the specified ground truth, and thus we can handle thousands of sliding windows in a short time.

Note that we use a method proposed in~\cite{Kai2001The} to judge whether a point is inside a polygon, and this method is also known as the crossing number algorithm or the even-odd rule algorithm~\cite{Galetzka2012A}. The comparison between previous method and our algorithm is shown in Figure~\ref{fig:monte}, our method shows satisfactory performance for computing polygonal area in practice.

\subsection{Finely localize text with quadrangle}
The fine procedure focuses on using those sliding windows with higher overlapping area to tightly localize text.
Unlike horizontal rectangle that can be determined by two diagonal points, we need to predict the coordinates of four points to localize a quadrangle. However, simply using the 4 points to shape a quadrangle is prone to be self-contradictory, because the subjective annotation may make the network ambiguous to decide which is the first point.
% In order to avoid self-contradictory prediction,
% we first use a proposed sequential protocol to order four points of each ground truth.
Therefore, before training, it is essential to order 4 points in advance.
% After getting positive sliding windows from rough procedure, we

\begin{figure}[htb]
  \centering
  \centerline{\includegraphics[width=8.2cm]{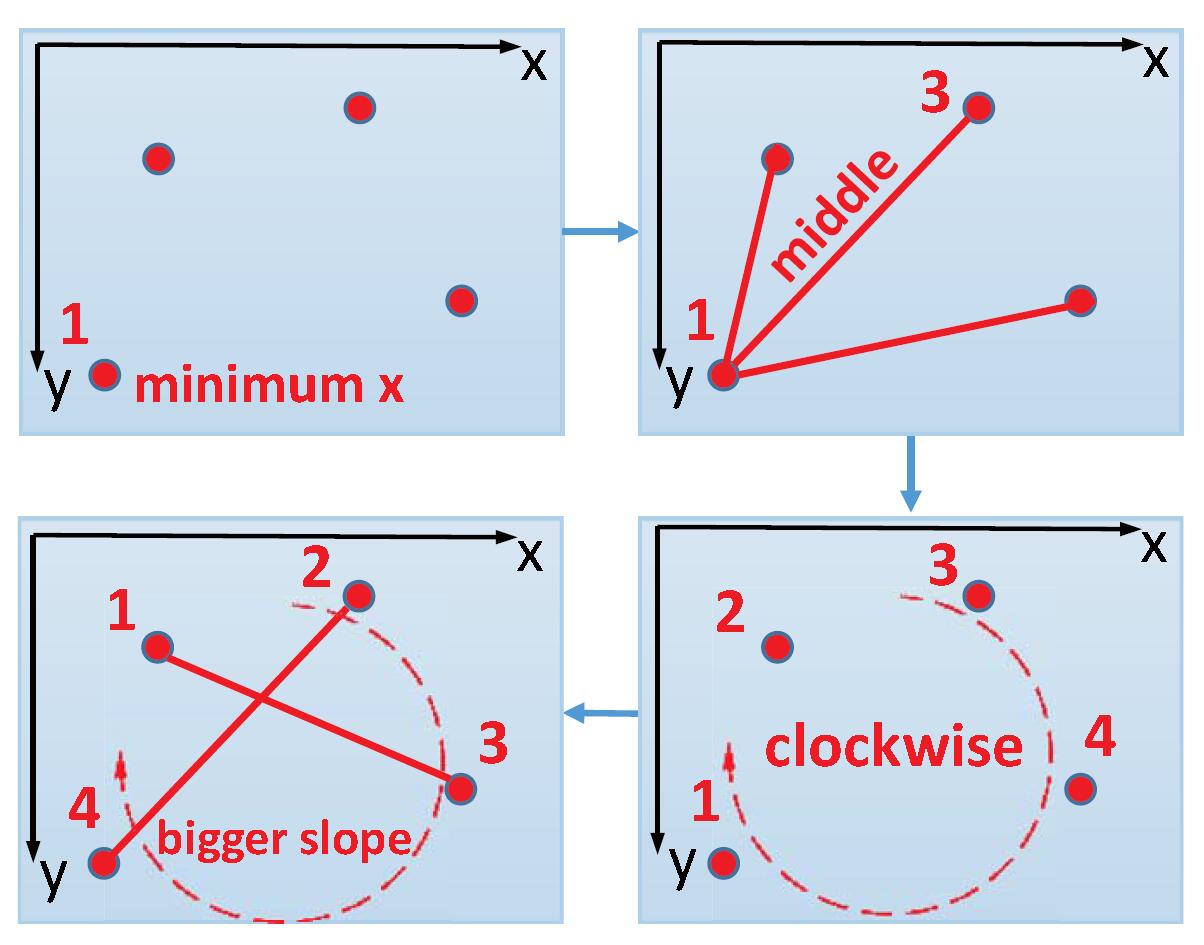}}
%  \vspace{2.0cm}
  \caption{Procedure of uniquely determining the sequence of four points from a plane convex quadrangle.}\label{fig:order}
\end{figure}
\textbf{Sequential protocol of coordinates.} The propose protocol can be used to determine the sequence of four points in the plane convex quadrangle, which contains four steps as shown in Figure~\ref{fig:order}.
% All ground truths from ICDAR 2015 dataset are labeled with 4 points.
First, we determine the first point with minimum value x. If two points simultaneously have the minimum x, then we choose the point with smaller value y as the first point. Second, we connect the first point to the other three points, and the third point can be found from the line with middle slope.
% The Third point can be determined from the middle segment.
The second and the fourth points are in the opposite side (defined ``bigger'' side and ``smaller'' side) of the middle line. Here, we assume middle line $L_{m}: ax+by+c=0,$
and we define an undetermined point $P(x_{p},y_{p})$. If $L_{m}(P)>0$, we assume $P$ is in the ``bigger'' side. If $L_{m}(P)<0$, $P$ is assumed in the ``smaller'' side.
Based on this assumption, the point in the ``bigger'' side would be assigned as second point, and the last point would be regarded as the fourth point.
% This step just like ordering them in clockwise began with the first point as shown in the third procedure of figure~\ref{fig:order}!!!!.
The last step is to compare the slopes between two diagonals ($line_{13}$ and $line_{24}$). From the line with bigger slope, we choose the point with smaller x as the new first point. Specially, if the bigger slope is infinite, the point that has smaller y would be chosen as the first point. Similarly, we find out the third point, and then the second and fourth point can be determined again. After finishing these four steps, the final sequence of the four points from a given convex quadrangle can be uniquely determined.

\begin{figure}[htb]
  \centering
  \centerline{\includegraphics[width=8.2cm]{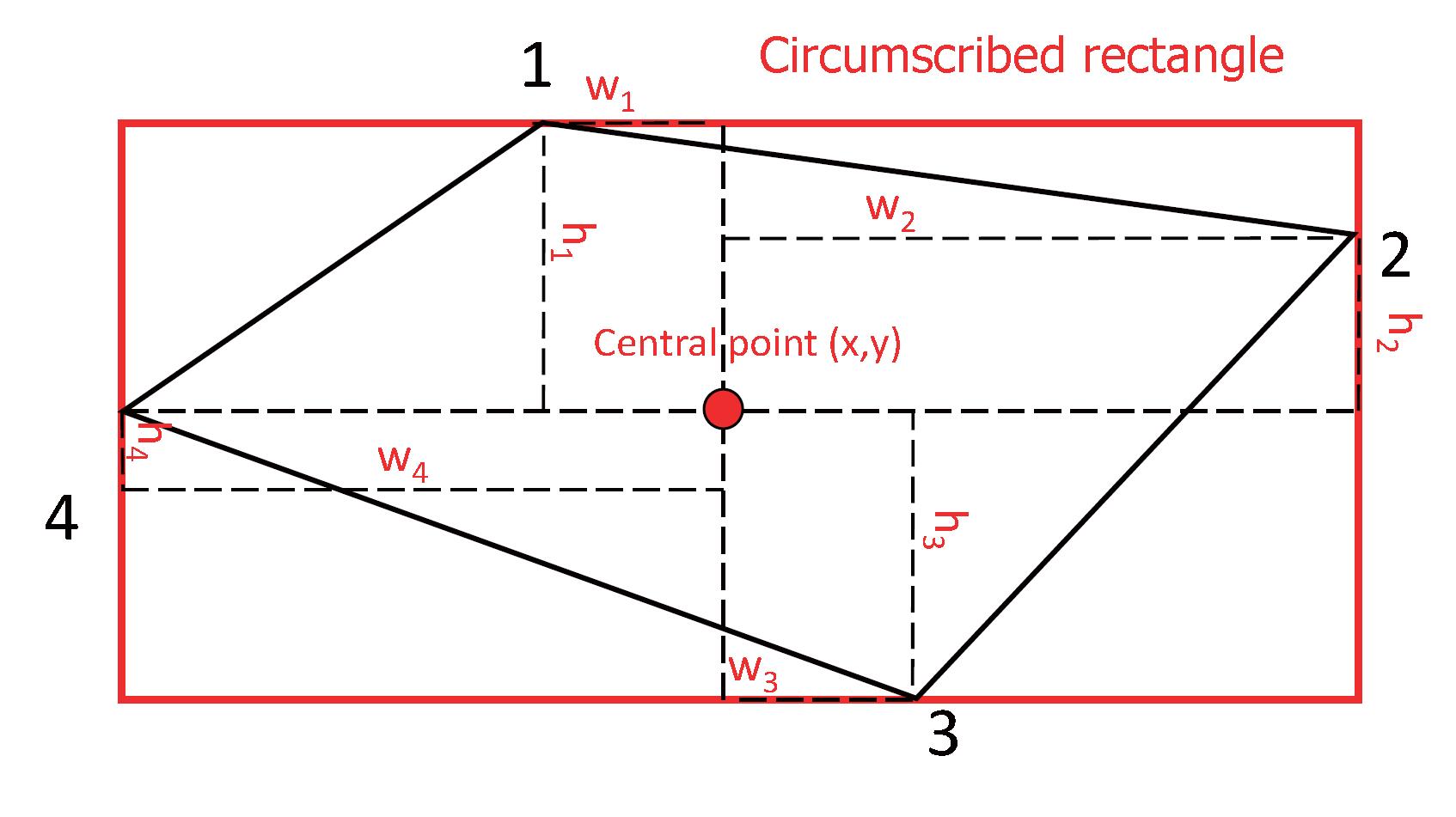}}
%  \vspace{2.0cm}
  \caption{ The position of each point of quadrangle can be calculated by central point and the relative lengths.
  % Different decisions made by humans and CGDL on ambiguous handwritten data. This figure suggests that even if the data are mislabeled by CGDL, the decisions can be deliberated. In addition, CGDL can help humans figure out wrong labels and correct these labels.
  }\label{fig:regress}
\end{figure}

Based on the sequential protocol,  DMPNet can clearly learn and regress the coordinate of each point by computing the relative position to the central point. Different from~\cite{Ren2016Faster} which regress two coordinates and two lengths for a rectangular prediction, our regressive method predicts two coordinates and eight lengths for a quadrilateral detection. For each ground truth, the coordinates of four points would be reformatted to $(x,y,w_{1},h_{1},w_{2},h_{2},w_{3},h_{3},w_{4},h_{4})$, where $x$, $y$ are the central coordinate of the minimum circumscribed horizontal rectangle, and $w_{i}, h_{i}$ are the relative position of the $i$-th point ($i = \{1,2,3,4\}$) to the central point. As Figure~\ref{fig:regress} shows, the coordinates of four points ($x_{1}$,$y_{1}$,$x_{2}$,$y_{2}$,$x_{3}$,$y_{3}$,$x_{4}$,$y_{4}$)$=$ ($x+w_{1}$,$y+h_{1}$,$x+w_{2}$,$y+h_{2}$,$x+w_{3}$,$y+h_{3}$,$x+w_{4}$,$y+h_{4}$). Note that $w_{i}$ and $h_{i}$ can be negative. Actually, eight coordinates are enough to determine the position of a quadrangle, and the reason why we use ten coordinates is because we can avoid regressing 8 coordinates, which do not contain relative information and it is more difficult to learn in practice~\cite{girshick2015fast}. Inspired by~\cite{Ren2016Faster}, we also use Lreg($p_{i}$;$p_{i}^{*}$) = R($p_{i}$-$p_{i}^{*}$) for multi-task loss, where R is our proposed loss function (smooth $Ln$) that would be described in section 3.4. $p^{*}$ $=$ $(p_{x}^{*},p_{y}^{*},p_{w1}^{*},p_{h1}^{*},p_{w2}^{*},p_{h2}^{*},p_{w3}^{*},p_{h3}^{*},p_{w4}^{*},p_{h4}^{*})$ represents the ten parameterized coordinates of the predicted bounding box, and $p$ $=$ $(p_{x},p_{y},p_{w1},p_{h1},p_{w2},p_{h2},p_{w3},p_{h3},p_{w4},p_{h4})$ represents the ground truth.
% $p = (p_{x},p_{y},p_{w1},p_{h1},p_{w2},p_{h2},p_{w3},p_{h3},p_{w4},p_{h4})$. $x_{i} = x+p_{wi}$
% the relationship can be showed in figure~\ref{fig:regress}
 % $x_{1} = x+p_{w1}$,$y_{1} = y+p_{h1}$,$x_{2} = x+p_{w2}$,$y_{2} = y+p_{h2}$,$x_{3} = x+p_{w3}$,$y_{3} = y+p_{h3}$,$x_{4} = x+p_{w4}$,$y_{4} = y+p_{h4}$.
% By the way, $p_{wi}$ and $p_{hi}$ $i={1,2,3,4}$ can be both positive and negative.

From the given coordinates, we can calculate the minimum x ($x_{min}$) and maximum x ($x_{max}$) of the circumscribed rectangle, and the width of circumscribed horizontal rectangle $w_{chr} = x_{max} -x_{min}$. Similarly, we can get the height $h_{chr}=y_{max} -y_{min}$.

% \begin{figure*}[htb]
%   \centering
%   \centerline{\includegraphics[width=17cm]{monte3.eps}}
% %  \vspace{2.0cm}
%   \caption{Comparison between previous method and our method in terms of computing overlapping area.}\label{fig:monte}
% \end{figure*}

We adopt the parameterizations of the 10 coordinates as following:

$d_{x}=\frac{p_{x}^{*}-p_{x}}{w_{chr}}$, $d_{y}=\frac{p_{y}^{*}-p_{y}}{h_{chr}}$, $d_{w_{1}}=\frac{p_{w_{1}}^{*}-p_{w_{1}}}{w_{chr}}$, $d_{h_{1}}=\frac{p_{h_{1}}^{*}-p_{h_{1}}}{h_{chr}}$,
$d_{w_{2}}=\frac{p_{w_{2}}^{*}-p_{w_{2}}}{w_{chr}}$, $d_{h_{2}}=\frac{p_{h_{2}}^{*}-p_{h_{2}}}{h_{chr}}$,
$d_{w_{3}}=\frac{p_{w_{3}}^{*}-p_{w_{3}}}{w_{chr}}$, $d_{h_{3}}=\frac{p_{h_{3}}^{*}-p_{h_{3}}}{h_{chr}}$,
$d_{w_{4}}=\frac{p_{w_{4}}^{*}-p_{w_{4}}}{w_{chr}}$, $d_{h_{4}}=\frac{p_{h_{4}}^{*}-p_{h_{4}}}{h_{chr}}$.
% $d_{w1}=p_{w1}/w_{chr}$,$d_{w1}=p_{w1}/w_{chr}$, $d_{w1}=p_{w1}/w_{chr}$, $d_{w1}=p_{w1}/w_{chr}$, $d_{h1}=p_{h1}/h_{chr}$, $d_{w2}=p_{w2}/w_{chr}$, $d_{h2}=p_{h2}/h_{chr}$, $d_{w3}=p_{w3}/w_{chr}$, $d_{h3}=p_{h3}/h_{chr}$, $d_{w4}=p_{w4}/w_{chr}$, $d_{h4}=p_{h4}/h_{chr}$,
% where x, xa, and x* are for the predicted box, anchor box, and ground-truth box respectively.
This can be thought of as fine regression from an quadrilateral sliding window to a nearby ground-truth box.
% Each regressor is responsible for one scale and one aspect ratio, and the k regressors do not share weights.
% As such, it is still possible to predict boxes of various sizes even though the features are of a fixed size/scale.
% Said this is the finely regress procedure.

%
% As the figure~\ref{fig:monte} shown, . Note that each
% Our method consists of three steps:
% we proposed a shared Monte-Carlo method to solve these issue as shown in figure [bababa]. Our computational method contain two parts.
% \begin{itemize}
%   \item
%   \item  If the circumscribed rectangle of each sliding window and the circumscribed rectangle of each ground truth do not have a intersection, the overlapping area is zero and we do not need to further compute.
%   \item  If the intersection is
%   Proposed a protocol that can uniquely determine the order of 4 points of arbitrary plane convex quadrangle. Based on the protocol, our method is able to use quadrangle to predict scene text, significantly reduce the background inference.
% \end{itemize}
%
% The figure [] shows our method. and compared the accuracy to previous method. we find our method has much more accuracy.
%

\begin{figure*}[htb]
\begin{minipage}[b]{.48\linewidth}
  \centering
  \centerline{\includegraphics[width = 8.25cm]{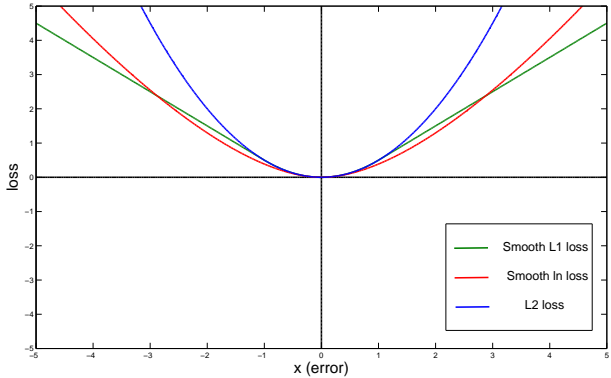}}
%  \vspace{1.5cm}
  \centerline{(a) forward loss functions. }\medskip
\end{minipage}
\hfill
\begin{minipage}[b]{0.48\linewidth}
  \centering
  \centerline{\includegraphics[width = 8.25cm]{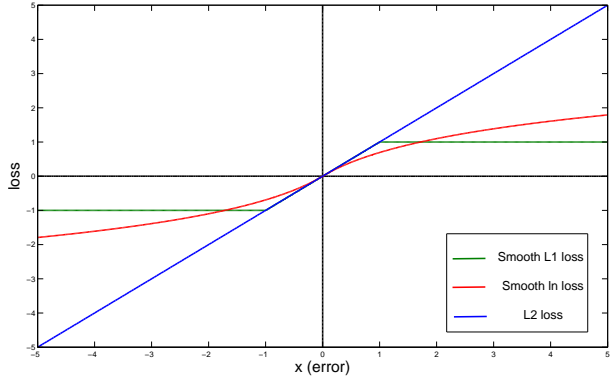}}
%  \vspace{1.5cm}
  \centerline{(b) backward deviation functions. }\medskip
\end{minipage}
\caption{Visualization of differences among three loss functions ($L2$, smooth $L1$ and smooth $Ln$). Here, the $L2$ function uses the same coefficient 0.5 with smooth $L1$ loss.   }\label{fig:loss}
\end{figure*}

\subsection{Smooth Ln loss}
% In our roughly recall procedure, the positive prior box is selected from the detection bounding box that has intersection over union (IoU) overlap greater than a overlapping threshold. The final prediction produce based on the positive prior box, than the error would be computed in the subsequence finely localize procedure. However,
Different from ~\cite{liu2015ssd,Ren2016Faster}, our approach uses a proposed smooth $Ln$ loss instead of smooth $L_{1}$ loss to further localize scene text. Smooth $L_{1}$ loss is less sensitive to outliers than the $L_{2}$ loss used in R-CNN~\cite{girshick2014rich}, however, this loss is not stable enough for adjustment of a data, which means the regression line may jump a large amount for small adjustment or just a little modification was used for big adjustment.
As for proposed smooth $Ln$ loss, the regressive parameters are continuous functions of the data, which means for any small adjustment of a data point, the regression line will always move only slightly, improving the precision in localizing small text. For bigger adjustment, the regression can always move to a moderate step based on smooth $Ln$ loss, which can accelerate the converse of training procedure in practice.
% during the training procedure, we always need higher adjustment to update the parameters of DMPNet, and smooth $Ln$ loss can provide much
% . where, instability property is explained by wikipedia, [describe in eleventh month], (the least squares solution  is stable in that, for any small adjustment of a data point, the regression line will always move only slightly; that is, the regression parameters are continuous functions of the data. for least absolute deviations loss, for a small horizontal adjustment of a datum, the regression line may jump a large amount.)
% Because the variance of scale of scene text, some text may fill the whole image, while some text is just 30 pixels width and 28 pixels height in a 1280x720 image. During our experiments, if we use L1 loss function, it is difficult to localize the big objects with small loss weight. Conversely, big weights is always hard to localize the very small objects
As mentioned in section 3.2, the recursive loss, $Lreg$, is defined over a tuple of true bounding-box regression targets $p^{*}$ and a predicted tuple $p$ for class text. The Smooth $L_{1}$ loss proposed in~\cite{girshick2015fast} is given by:
% $p^{*} = (p_{x}^{*},p_{y}^{*},p_{w1}^{*},p_{h1}^{*},p_{w2}^{*},p_{h2}^{*},p_{w3}^{*},p_{h3}^{*},p_{w4}^{*},p_{h4}^{*})$,
% $p = (p_{x},p_{y},p_{w1},p_{h1},p_{w2},p_{h2},p_{w3},p_{h3},p_{w4},p_{h4})$.
% 10 value need to be predicted, in fact, 8 is enough to represent the exact position, but use 10, it is more easy to operate. The set of these ten value is $S=\{x,y,w1,h1,w2,h2,w3,h3,w4,h4\}$.

% Based on our training experiments, Smooth $L_{1}$
\begin{equation}\label{equation:1}
  Lreg(p;p^{*}) = \sum_{i\in S} smooth_{L_{1}}(p_{i},p^{*}),
\end{equation}
in which,
\begin{equation}\label{equation:2}
  smooth_{L_{1}}(x) = \left\{
                        \begin{array}{ll}
                          0.5x^{2} & \hbox{if $|x| < 1$} \\
                          |x|-0.5 & \hbox{otherwise.}
                        \end{array}
                      \right.
\end{equation}
The x in the function represents the error between predicted value and ground truth ($x = w\cdot (p-p_{*})$). The deviation function of $smooth_{L_{1}}$ is:
\begin{equation}\label{equation:3}
  deviation_{L_{1}}(x) = \left\{
                        \begin{array}{ll}
                          x & \hbox{if $|x| < 1$} \\
                          sign(x) & \hbox{otherwise.}
                        \end{array}
                      \right.
\end{equation}
As equation~\ref{equation:3} shows, the deviation function is a piecewise function while the smooth $Ln$ loss is a continuous derivable function.
% The smooth $L_{2}$ loss is a simple
% Equation \ref{equation:3} shows
The proposed Smooth $Ln$ loss is given by:
\begin{equation}\label{equation:4}
  Lreg(p;p^{*}) = \sum_{i\in S} smooth_{Ln}(p_{i},p^{*}),
\end{equation}
in which,
\begin{equation}\label{equation:5}
  smooth_{Ln}(x) = (|d|+1)ln(|d|+1)-|d|,
\end{equation}
and the deviation function of $smooth_{Ln}$ is:
\begin{equation}\label{equation:6}
  deviation_{Ln}(x) =  sign(x)\cdot ln(sign(x)\cdot x +1).
\end{equation}
Equation~\ref{equation:5} and equation~\ref{equation:6} are both continuous function with a single equation. For equation~\ref{equation:6}, it is easy to prove $|x|\geq |deviation_{Ln}(x)|$, which means the smooth $Ln$ loss is also less sensitive to outliers than the $L_{2}$ loss used in R-CNN~\cite{girshick2014rich}. A intuitive representation of the differences among three loss functions is shown in Figure~\ref{fig:loss}. The comparisons of properties in terms of robustness and stability are summarized in Table~\ref{tb:1}. The results demonstrate that the smooth $Ln$ loss promises better text localization and relatively tighter bounding boxes around the texts.

% \begin{equation}\label{equation:7}
% sign(x)\cdot ln(sign(x)\cdot x +1)\  for\  \forall x\in R,
% \end{equation}

% Smooth $ln$ loss incentives better text localization, and easier to return tight bounding boxes around objects, which is given by:
% \begin{equation}\label{4}
  % L_{loc}(t,t^{*}) = /sum_{x,y,w1,h1,w2,h2,w3,h3,w4,h4}
% \end{equation}

% The total different between these loss can be listed in the table ()
\begin{table}
\begin{center}
\footnotesize
\begin{tabular}{|c|c|c|c|}
\hline
property & $L2$ loss  & smooth $L_{1}$ loss  & smooth ${Ln}$ loss \\
\hline
Robustness & Worst & Best & Good\\
Stability & Good & Worst & Best\\
\hline
\end{tabular}
\end{center}
\caption{Different properties of different loss functions. \textbf{Robustness} represents the ability of resistance to outliers in the data and \textbf{stability} represents the capability of adjusting regressive step.}\label{tb:1}
\end{table}

% layer is then utilize those  These the prior box
\section{Experiments}
Our testing environment is a desktop running Ubuntu 14.04 64bit version with TitanX. In this section, we quantitatively evaluate our method on the public dataset: ICDAR 2015 Competition Challenge 4: ``Incidental Scene Text''~\cite{Karatzas2015ICDAR}, and as far as we know, this is the only one dataset in which texts are both word-level and multi-oriented. All results of our methods are evaluated from its online evaluation system, which would calculate the recall rate, precision and F-measure to rank the submitted methods. The general criteria of these three index can be explained below:
\begin{itemize}
  \item \textbf{Recall rate} evaluates the ability of finding text.
  \item \textbf{Precision} evaluates the reliability of predicted bounding box.
  \item \textbf{F-measure} is the harmonic mean (Hmean) of recall rate and precision, which is always used for ranking the methods.
\end{itemize}

Particularly, we simply use official 1000 training images as our training set without any extra data augmentation, but we have modified some rectangular labels to quadrilateral labels for adapting to our method.
% he experiments are conducted in Ubuntu 14.04 with TitianX gpu.
% We run our experiments in Ubuntu 14.04 64bit version with TitanX gpu. The result is evaluated on the online evaluation system, which compute the real overlap instead of approximate one between predicted value and ground truth.
% We quantitatively evaluate the proposed method on the most widely used and most difficult public datasets: ICDAR 2015 Competition Challenge 4: ``Incidental Scene Text'' [refers],

 % We quantitatively evaluate the proposed method on the most widely used and most difficult public datasets: ICDAR 2015 Because ICDAR 2015 is the only one that. Particularly, we simply use the image of from it's public dataset.
% we use ICDAR 2015 online test system to evaluate our method,  The test system would give us the result of word-level recall rate, precision and f measure.

% \subsection{Datasets}
% We evaluate our method on the two word-level dataset, one is ICDAR 2015 Competition Challenge 4 which is the only one public dataset based on both word-level and multi-oriented.
% The other is ICDAR 2013 that is a horizontal text dataset.
% All training set are

\textbf{Dataset - ICDAR 2015 Competition Challenge 4 ``Incidental Scene Text''~\cite{Karatzas2015ICDAR}.} Different from the previous ICDAR competition, in which the text are well-captured, horizontal, and typically centered in images. The datasets includes 1000 training images and 500 testing incidental scene image in where text may appear in any orientation and any location with small size or low resolution and the annotations of all bounding boxes are marked at the word level.
 % The evaluation protocol of this dataset inherits from [refer]. Note that this competition provides an online evaluation system and our method is evaluated in the same way.

 % To satisfy the requirement of ICDAR 2015 measurement, we perform the word partition on the text lines generated by our method according to the blanks between words.

% \textbf{ICDAR 2013 Competition Challenge 2 - Focus Scene Text dataset.} The ICDAR 2013 dataset is a horizontal text database which is used in previous ICDAR competitions. This dataset consists of 229 images for training and 233 images for testing. The evaluation algorithm is introduced by [refer] and we evaluate our method on the ICDAR2013 online evaluation system. Since this dataset also provides wordlevel annotations, we adopt the same word partition procedure as we did on ICDAR 2015 dataset.

% \subsection{Implementation details}
\textbf{Baseline network.} The main structure of DMPNet is based on the VGG-16 model~\cite{simonyan2014very},
 % in which we converted the full connected 6 (fc6) and full connected 7 (fc7) layers to convolutional layers.
Similar to Single Shot Detector~\cite{liu2015ssd}, we use the same intermediate convolutional layers to apply quadrilateral sliding windows. All input images would be resized to a 800x800 for preserving tiny texts.
% We only use its public dataset as training set to trained our method. Our method is based on the VGG-16 model. The structure is similar to SSD [refer],

% \subsection{Experimental results}
\textbf{Experimental results.}  For comprehensively evaluating our algorithm, we collect and list the competition results~\cite{Karatzas2015ICDAR} in Table~\ref{tab:2}. The previous best method of this dataset, proposed by Yao~\etal~\cite{Yao2015Incidental}, achieved a F measure of 63.76\% while our approach obtains 70.64\%. The precision of these two methods are comparable but the recall rate of our method has greatly increased, which is mainly due to the quadrilateral sliding windows described in section 3.1.
% We collect competition results~\cite{Karatzas2015ICDAR} as listed in Tab. 2 for comprehensive comparisons. Our method achieves the best F-measure over all methods.
% RESULT!!!!!!!
% \textbf{ICDAR 2015 Competition Challenge 4 - Incidental Scene Text dataset.}As this dataset has been released recently for the competition in ICDAR2015, there is no literature to report the experimental result on it. Therefore, we collect competition results~\cite{Karatzas2015ICDAR} as listed in Tab. 2 for comprehensive comparisons. Our method achieves the best F-measure over all methods.

\begin{table}[!t]
\caption{Evaluation on the ICDAR 2015 competition on robust reading challenge 4 ``Incidental Scene Text'' localization.}% in the permutation invariant setting
\label{tab:2}
\centering
\scriptsize
\begin{tabular}{|c|c|c|c|}
  \hline
  Algorithm & Recall (\%)  & Precision (\%) & Hmean (\%) \\
  \hline
  \hline
   \textbf{Baseline (SSD-VGGNet)} & 25.48 & 63.25 & 36.326 \\
   \hline
  \textbf{Proposed DMPNet}  & \textbf{68.22} & 73.23 & \textbf{70.64} \\
  \hline
  Megvii-Image++~\cite{Yao2015Incidental} & 56.96 & 72.40 & 63.76 \\
  \hline
  CTPN~\cite{Tian2016Detecting} & 51.56 & 74.22 & 60.85 \\
  \hline
  MCLAB\_FCN~\cite{Karatzas2015ICDAR} & 43.09 & 70.81 & 53.58 \\
  \hline
  StardVision-2~\cite{Karatzas2015ICDAR} & 36.74 & \textbf{77.46} & 49.84\\
  \hline
  StardVision-1~\cite{Karatzas2015ICDAR}  & 46.27 & 53.39  & 49.57\\
  \hline
  CASIA\_USTB-Cascaded~\cite{Karatzas2015ICDAR} & 39.53 & 61.68 & 48.18 \\
  \hline
  NJU\_Text~\cite{Karatzas2015ICDAR}   & 35.82 & 72.73 & 48.00\\
  \hline
  AJOU~\cite{Koo2013Scene}  & 46.94 & 47.26 & 47.10 \\
  \hline
  HUST\_MCLAB~\cite{Karatzas2015ICDAR}  & 37.79 & 44.00 & 40.66\\
  \hline
  Deep2Text-MO~\cite{Yin2015Multi}  & 32.11 & 49.59 & 38.98\\
  \hline
  CNN Proposal~\cite{Karatzas2015ICDAR}  & 34.42 & 34.71 & 34.57 \\
  \hline
  TextCatcher-2~\cite{Karatzas2015ICDAR}  & 34.81 & 24.91 & 29.04 \\
  \hline
\end{tabular}
\end{table}

\begin{figure}[htb]
  \centering
  \centerline{\includegraphics[width=8.25cm]{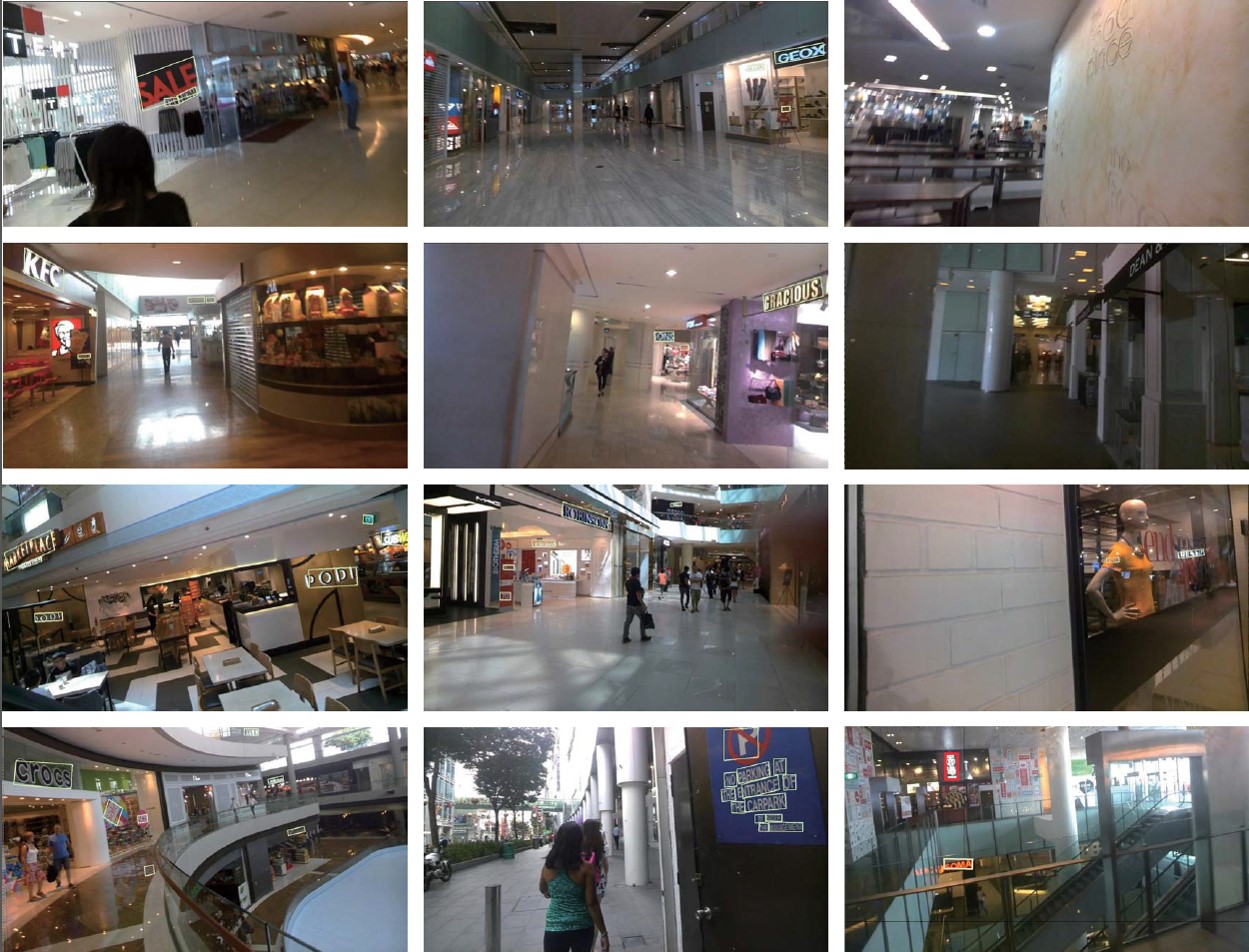}}
%  \vspace{2.0cm}
  \caption{Experimental results of samples on ICDAR 2015 Challenge 4, including multi-scale and multi-language word-level text. Our method can tightly localize text with less background information as shown in the first two columns. Top three images from last column are the failure recalling cases of the proposed method. Specially, some labels are missed in some images, which may reduce our accuracy as the red bounding box listed in the fourth image of the last column.  }\label{fig:visual}
\end{figure}

Figure~\ref{fig:visual} shows several detected results taken from the test set of ICDAR 2015 challenge 4. DMPNet can robustly localize all kinds of scene text with less background noise. However, due to the complexity of incidental scene, some false detections still exist, and our method may fail to recall some inconspicuous text as shown in the last column of Figure~\ref{fig:visual}.
% The detecting results are shown in the figure~\ref{fig:visual}

% \begin{figure*}[htb]
%   \centering
%   \centerline{\includegraphics[width=17cm]{final2-5.eps}}
% %  \vspace{2.0cm}
%   \caption{Experimental results of samples on ICDAR 2015 Challenge 4, including multi-scale and multi-language word-level text. Green are ground truth given by online evaluatOur method can tightly localize severely distorted samples }\label{final2}
% \end{figure*}

\section{Conclusion and future work}
In this paper, we have proposed an CNN based method, named Deep Matching Prior Network (DMPNet), that can effectively reduce the background interference. The DMPNet is the first attempt to adopt quadrilateral sliding windows, which are designed based on the priori knowledge of textual intrinsic shape, to roughly recall text. And we use a proposed sequential protocol and a relative regressive method to finely localize text without self-contradictory. Due to the requirement of computing numerous polygonal overlapping area in the rough procedure, we proposed  a shared Monte-Carlo method for fast and accurate calculation. In addition, a new smooth $Ln$ loss is used for further adjusting the prediction, which shows better overall performance than $L2$ loss and smooth $L1$ loss in terms of robustness and stability. Experiments on the well-known ICDAR 2015 robust reading challenge 4 dataset demonstrate that DMPNet can achieve state-of-the-art performance in detecting incidental scene text. In the following, we discuss an issue related to our approach and briefly describe our future work.

 \textbf{Ground truth of the text.} Texts in camera captured images are always with perspective distortion. However rectangular constraints of labeling data may bring a lot of background noise, and it may lose information for not containing all texts when labeling marginal text. As far as we know, ICDAR 2015 Challenge 4 is the first dataset to use quadrilateral labeling, and our method prove the effectiveness of utilizing quadrilateral labeling. Thus, quadrilateral labeling for scene text may be more reasonable.
% is intuitive This robust algorithm
% During the rough procedure, for each ground truth, we need to compute its overlapping area with every quadrilateral sliding window

\textbf{Future Work.} The high recall rate of the DMPNet mainly depends on numerous prior-designed quadrilateral sliding windows. Although our method have been proved effective, the man-made shape of sliding window may not be the optimal designs. In future, we will explore using shape-adaptive sliding windows toward tighter scene text detection.

{\small
% \bibliographystyle{ieee}
% \bibliography{egbib}
\bibliographystyle{model2-names}

\end{document}